\documentclass[10pt, a4paper]{article}

\usepackage[final]{lrec2026} 
\usepackage{booktabs}
\usepackage{multirow}
\title{Decoding News Narratives: A Critical Analysis of Large Language Models in Framing Detection}

\name{Valeria Pastorino,   Jasivan Alex Sivakumar,   Nafise Sadat Moosavi} 

\address{Department of Computer Science \\ University of Sheffield \\ United Kingdom \\
         \{vpastorino1|jasivakumar1|n.s.moosavi\}@sheffield.ac.uk\\}

\abstract{
The growing complexity and diversity of news coverage have made framing analysis a crucial yet challenging task in computational social science. Traditional approaches, including manual annotation and fine-tuned models, remain limited by high annotation costs, domain specificity, and inconsistent generalisation. Instruction-based large language models (LLMs) offer a promising alternative, yet their reliability for framing analysis remains insufficiently understood.
In this paper, we conduct a systematic evaluation of several LLMs, including GPT-3.5/4, FLAN-T5, and Llama 3, across zero-shot, few-shot, and explanation-based prompting settings. Focusing on domain shift and inherent annotation ambiguity, we show that model performance is highly sensitive to prompt design and prone to systematic errors on ambiguous cases. Although LLMs, particularly GPT-4, exhibit stronger cross-domain generalisation, they also display systematic biases, most notably a tendency to conflate emotional language with framing.
To enable principled evaluation under real-world topic diversity, we introduce a new dataset of out-of-domain news headlines covering diverse subjects. Finally, by analysing agreement patterns across multiple models on existing framing datasets, we demonstrate that cross-model consensus provides a useful signal for identifying contested annotations, offering a practical approach to dataset auditing in low-resource settings.
 \\ \newline  }

\begin{document}

\maketitleabstract

\section{Introduction}

\label{sec:intro}
In today’s digital age, the rapid growth of news sources and the widespread dissemination of information have intensified the need for unbiased and transparent reporting. At the same time, news coverage is often shaped through framing, a communication strategy that selectively emphasizes certain aspects of an issue in order to influence public perception and promote particular interpretations \cite{linglinguag18941}. As a result, framing represents a persistent obstacle to maintaining a well-informed public audience. It affects not only how events are perceived and remembered, but also how they are evaluated, discussed, and translated into policy preferences. For instance, a government policy change may be framed as ``Government’s heartless cutbacks leave thousands without essential services'' or, alternatively, as ``Government announces reduction in funding for public services''.

\begin{figure}
    \centering
    \includegraphics[width=0.8\linewidth]{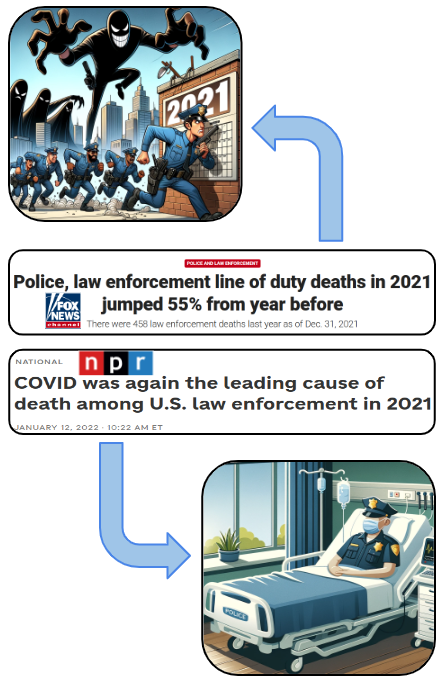}
    \caption{Example of two different ways of framing the same news.}
    \label{fig:frame}
\end{figure}

In social science, framing is understood as a mechanism for guiding audience interpretation through selective emphasis \cite{goffman1974frame,entman_framing_1993}. Building on this tradition, we consider a headline to be framed when it selectively emphasizes certain aspects of an event while downplaying others in order to steer readers toward a particular interpretation. This definition distinguishes framing from merely emotional or descriptive language and grounds our computational task in established communication theory.

Despite its importance, empirical studies of framing have traditionally relied on small, manually annotated datasets \cite{baumer_testing_2015}. While supervised computational approaches have been proposed to scale such analyses, their performance often deteriorates under domain shift, limiting their applicability across diverse news contexts \citep{sinelnik-hovy-2024-narratives}. As online media continues to expand in volume and scope, there is a growing need for scalable and reliable methods for framing analysis.
Large language models (LLMs) offer a promising alternative to traditional approaches. Instruction-following, pre-trained LLMs can be adapted to new tasks with minimal supervision, reducing the cost of domain-specific annotation. However, despite their increasing use in social-science research, their reliability for detecting nuanced framing remains insufficiently understood.

This paper addresses this gap through a systematic investigation of instruction-based LLMs for news framing detection. We evaluate GPT-4, GPT-3.5 Turbo, Llama~3, and FLAN-T5 across zero-shot, few-shot, and explanation-based prompting settings, with an emphasis on robustness, bias, and failure modes. To enable a principled analysis of cross-domain generalisation, we first introduce a new dataset of real-world news headlines spanning diverse topics. Using this resource, we examine how reliably these models detect framing under domain shift and limited supervision, how their predictions vary with prompt design, and what systematic biases arise in practice.

Our analysis shows that although LLMs, especially GPT-4, exhibit stronger cross-domain generalisation than fine-tuned models, their predictions remain highly sensitive to prompt design and prone to systematic errors on inherently ambiguous cases. In particular, we identify a consistent tendency of GPT-4 to conflate emotional language with framing, leading to recurrent false positives. Finally, by analysing patterns of agreement and disagreement across models on existing framing datasets, we show that cross-model consensus provides a useful signal for identifying contested or potentially problematic annotations, offering a practical tool for dataset auditing.\footnote{
For reproducibility and future research, our dataset is available \url{https://github.com/vpastorino/ITW-dataset}.}

\section{Related Work}

\subsection{Automatic Framing Detection}
A wide range of NLP methods has been proposed for automatic framing detection. Early studies primarily relied on topic modeling approaches, including Topic Modeling, Structural Topic Modeling, and Hierarchical Topic Modeling, to uncover themes in large document collections \cite{DIMAGGIO2013570, nguyen-etal-2015-tea, gilardi2021}. While effective for identifying what is discussed, these methods often provide limited insight into how issues are framed. 

Latent Dirichlet Allocation (LDA) Topic Modeling, for instance, served as a starting point for creating lists of frames deductively in tools like the one presented by \citet{bhatia_openframing_2021} for computational framing analysis. However, as noted by \citet{ali_survey_2022}, the emphasis in such approaches remains on detecting topics rather than the nuanced framing of those events. The focus on topics could also derive from the connections between agenda setting and framing strategies in computational social sciences, with studies analysing these two phenomena together \cite{field_framing_2018}.

Further analyses have included pragmatics cues, examining how specific word choices, like the use of ``again'' in ``Again, Dozens of Refugees Drowned'', subtly influence reader perception \cite{yu_again_2022}. 
This shift towards granular analysis is complemented by advanced models, including Neural Network and deep learning techniques, which offer refined tools for detecting framing nuances \cite{burscher2016, card_media_2015, liu_detecting_2019, mendelsohn_modeling_2021}.
\citet{tourni_detecting_2021} demonstrated that combining transformer models for processing news headlines with residual network models to process news lead images could improve the accuracy of framing detection. Similarly, \citet{classifying_2017} explored the use of various neural networks, such as LSTMs, BiLSTMs, and GRUs, for frame prediction at the sentence level using the Media Frame Corpus (MFC) \cite{card_media_2015}. 

Building on this foundation, \citet{liu_detecting_2019} and \citet{akyurek_multi-label_2020} fine-tuned BERT \cite{devlin2019bert} to predict frames in news headlines. Their work resulted in the creation of the Gun Violence Frame Corpus (GVFC), a benchmark dataset for framing analysis which will be further discussed in section \ref{datasets}. 

In contrast to the supervised methodologies outlined above, which often struggle with generalisation due to their reliance on domain-specific training data \cite{ali_survey_2022}, our work explores an alternative approach using instruction-following language models, which potentially offer a more flexible and scalable solution for detecting framing in a broad array of news contexts. 

\subsection{Datasets } \label{datasets}

\paragraph{Media Frame Corpus (MFC)}

MFC  \citep{card_media_2015} is a collection of annotated U.S. newspaper articles on topics like immigration, smoking, and same-sex marriage, analysed for framing. Utilising the Policy Frames Codebook (PFC) by \citet{Boydstun2014}, the MFC adopts 14 frame dimensions such as ``security and defense'' and ``cultural identity'' for categorising policy discourse. Despite achieving an inter-coder reliability (ICR) of 0.60, critiques, particularly \citet{ali_survey_2022}, argue that the PFC’s broad dimensions conflate topics with frames, potentially missing nuanced strategic framing.

Moreover, MFC categorises content into wide-ranging dimensions (i.e., politics, economic, etc.) that might not always precisely capture the specific framing intended by a news headline \cite{ali_survey_2022}. This categorisation can make it difficult to directly identify if and how a headline is framed without a deeper, nuanced analysis. 

\paragraph{Gun Violence Frame Corpus (GVFC)}
Another significant dataset in the field of framing analysis is the GVFC, introduced by \citet{liu_detecting_2019}. This dataset concentrates on the issue of Gun Violence in the U.S. The creation process began with defining nine distinct ``frames'' related to the topic, drawing from existing literature and a preliminary data analysis. A specialised codebook was then developed, serving as a training tool for annotators along with annotation guidelines. 

GVFC is made of 2990 news headlines, with 2616 headlines specific to the issue of Gun Violence in the United States. All the in-domain headlines are coded to have a primary frame, while only 319 have two frames.
For instance, the headline ``It’s Time to Hand the Mic to Gun Owners'' is annotated with ``Public opinion'' as the first frame and ``2nd Amendment'' as the second frame. Similarly, ``Trevor Noah: The Second Amendment Is Not Intended for Black People'' is annotated with ``2nd Amendment'' and ``Race/Ethnicity'' frames \citep{liu_detecting_2019}.

\paragraph{Non-English Data} 
Expanding the scope of framing analysis to non-English content, \citet{akyurek_multi-label_2020} introduced a multilingual extension of the Gun Violence Frame Corpus, which encompasses news headlines in German, Turkish, and Arabic, focusing on U.S. gun violence. This extension involved training two native speakers per language to annotate headlines - 350 in German, 200 in Turkish, and 210 in Arabic. 

\citet{piskorski_semeval-2023_2023} presented an annotated dataset made of articles spanning nine languages: English, French, German, Georgian, Greek, Italian, Polish, Russian, and Spanish. This dataset addresses a variety of topics including the COVID-19 pandemic, abortion-related legislation, migration, Russo-Ukrainian war, and various parliamentary elections. The annotation process made use of the PFC codebook, using the 15 dimensions as frames \cite{piskorski2023news}.

\subsection{Evaluating LLMs in Social Science}

Applying LLMs to social science tasks, such as evaluating sociability \cite{choi_llms_2023}, morality \cite{abdulhai_moral_2023}, and controversial issues and bias \cite{sun_delphi_2023}, has received increasing interest, showcasing a wide range of strengths and limitations of the examined language models unique to each task. This diversity stems from the specific challenges and nuances of social phenomena. Although LLMs excel in generating and understanding human-like text, the complex requirements of social science tasks necessitate a detailed, task-specific examination of their performance and reliability. 

In this work, we contribute to the expanding research on the applicability of LLMs in social sciences by specifically investigating their reliability in detecting framing.

\section{Experimental Setup}
\subsection{Data}
For our evaluation, we select GVFC \cite{liu_detecting_2019}, motivated by its comprehensive coverage of U.S. Gun Violence framing as well as the high ICR met in the annotation process. In our experiments, we have excluded the headlines that are not relevant to Gun Violence and hence, are not annotated with framing information. 
The dataset's annotations identify whether each headline reflects any of nine critical aspects of gun violence framing: gun rights, gun control, politics, mental health, public/school safety, race/ethnicity, public opinion, social/cultural issues, and economic consequences. We consider headlines tagged with any of the above categories as framed, and all others as not framed.
Further, we exclude 22 relevant headlines in order to use them for few-shot in domain prompting, leaving 2594 relevant headlines for our analysis. Of these, 1293 are framed and 1301 are not framed. Hence, the majority label is Not Framed and the majority baseline is 50.15\%.

\subsection{Models}
To evaluate the performance of contemporary large language models in framing detection, we consider two widely used closed-source models: GPT-4\footnote{GPT-4-0613. Temperature parameter set to 0.} \cite{openai2023gpt4} and GPT-3.5-Turbo\footnote{GPT-3.5-turbo-0613. Temperature parameter set to 0.} \cite{ye_comprehensive_2023}. These models are known for their strong performance across a wide range of NLP tasks and their ability to follow complex instructions. We select them due to their widespread adoption and practical accessibility for social science researchers, as they can be used via paid APIs without requiring specialised hardware or extensive technical setup.

In addition, we include two open-source alternatives: Llama~3 \cite{dubey2024llama} and FLAN-T5 \cite{flant5}, which have demonstrated competitive performance across diverse benchmarks \cite{chung_scaling_2022}. We evaluate the 8B version of Llama~3\footnote{Meta-Llama-3-8B-Instruct. Temperature parameter set to 0.1.} and multiple variants of FLAN-T5, including small (77M parameters), base (248M parameters), and large (783M parameters), to examine the effect of model scale on framing detection.

All models are evaluated using a unified set of prompts under three experimental conditions: (1) a zero-shot setting, assessing models’ baseline capabilities; (2) a few-shot setting, in which a small number of examples are provided; and (3) an explanation-based prompting setting, where models are asked to justify their predictions.

Our evaluation framework focuses on assessing model behaviour without task-specific fine-tuning, reflecting realistic conditions in social science research, where annotated datasets are often limited. This design allows us to examine the robustness and generalisability of LLMs for framing detection under low-supervision settings.

\subsection{Zero-Shot Prompting}
In the zero-shot setting, we evaluate model performance using two prompt variants. In the first, models are asked to determine whether a headline is framed without any additional context or examples.\footnote{GPT and Llama prompt: ``Decide whether this claim is framed''.\\FLAN-T5 prompt: ``Is this claim framed? OPTIONS Yes | No''. The FLAN-T5 prompt is designed to align with its instruction-based pre-training format.}

In the second variant, we augment the prompt with an explicit definition of framing to guide the model’s decision. Specifically, framing is defined as ``a communication strategy often used in journalism and political language, where certain aspects of an issue are highlighted while others are minimised or ignored, thereby promoting a particular interpretation of that issue''.\footnote{This definition is grounded in established accounts of framing \cite{goffman1974frame,entman_framing_1993,entman_framing_2007}.} Models are then asked to classify headlines according to this definition. This setting allows us to assess whether conceptual guidance improves framing detection.

\subsection{Few-Shot Prompting}
In the few-shot setting, we extend the zero-shot setup by providing a small number of labeled examples of framed and non-framed headlines within the prompt. These experiments are designed to examine how example-based supervision influences model behaviour and performance.

\paragraph{Impact of Example Quantity}
To examine how the number of examples affects model performance, we evaluate two few-shot configurations. The first includes a minimal set of two GVFC examples, one framed and one not framed, serving as a baseline for assessing the effect of example-based prompting. The second includes eight examples, of which four are framed, allowing us to analyse the impact of increased supervision.

\begin{table*}[!htb]
    \centering
    \footnotesize
    \begin{tabular}{ll|cccc|cccc|cccc} \toprule
     & & \multicolumn{4}{c|}{\bfseries GPT-3.5 Turbo} & \multicolumn{4}{c|}{\bfseries GPT-4} & \multicolumn{4}{c}{\bfseries Llama3 8B}  \\
      & & & & \multicolumn{2}{c|}{Explainable} & & & \multicolumn{2}{c}{Explainable} & & & \multicolumn{2}{c}{Explainable} \\
      & & F$_1$ & Acc. & F$_1$ & Acc. &  F$_1$ & Acc. &  F$_1$ & Acc. &  F$_1$ & Acc. &  F$_1$ & Acc. \\ \midrule
       \multirow{2}{*}{ZS}  & No Def.  & 20.48 & 52.70 & 61.43 & 53.28 & 64.96 & 58.40 & 63.75 & 60.41 & 51.73 & 57.19 & 58.59 & 55.81\\
         & +Def. & 51.55 & 59.14 & 59.03 & 58.79 & 65.36  & 63.84 & 64.64 & 60.99 & 52.66 & 53.55 & 57.42 & 50.35\\ 
         \midrule
         \multirow{5}{*}{FS}  
         & 2Ex. & 26.83 & 54.16 & \bfseries 65.84 & 61.68 & 58.25 & 65.84 & 64.23 & 64.92 & 57.23 & 59.08 & 56.93 & 56.18 \\
         & 8Focus. & 40.72 & 58.25 & 65.32 & 61.57 & 64.50 & 63.30 & 66.44 & 63.96 & 52.29 & 61.47 & 55.39 & 60.12 \\
         & 8Varied & 46.36 & 60.22 & 64.40 & 60.79 & \bfseries 68.51 & 65.38 & \bfseries 70.41 & \bfseries 66.92 & \textbf{58.58} & 61.94 & \textbf{61.66} & 59.08\\
         & 8Cross & 41.22 & 57.67 & 60.76 & 62.08 & 59.42 & \bfseries 66.04 & 59.09 & 64.61 & 47.84 & \textbf{63.05} & 53.40 & \textbf{60.49} \\
         & 8Mixed & \bfseries 56.93 & \bfseries 63.49 & 64.60 & \bfseries 62.99 & 63.33 & 64.37 & 63.87 & 63.76 & 58.32 & 62.21 & 57.98 & 59.72  \\
          \bottomrule
    \end{tabular}
    \caption{Comparative performance of GPT and Llama3 models showing ``zero-shot'' results with and without task definition, ``few-shot'' results with 2 or 8 examples, and ``Explainable'' results when models explain predictions. ``Acc.'' columns report overall accuracy, and ``F$_1$'' reports detection of framed headlines.}
    \label{tab:gpt}
\end{table*}

\begin{table*}[!htbp]
\centering
    \footnotesize
\begin{tabular}{c|cccc|cccc|cccc}
\toprule 
   & \multicolumn{4}{c|}{\bfseries GPT-3.5 Turbo} & \multicolumn{4}{c|}{\bfseries GPT-4} & \multicolumn{4}{c}{\bfseries Llama3 8B}\\
   & & & \multicolumn{2}{c|}{Explainable} & & & \multicolumn{2}{c|}{Explainable} & & & \multicolumn{2}{c}{Explainable}
   \\
   & F$_1$           & Acc.         & F$_1$          & Acc.        & F$_1$          & Acc.        & F$_1$         & Acc.    & F$_1$          & Acc.        & F$_1$         & Acc.          \\ \midrule
No Definition            & 20.48        & 52.70     & 61.43 &  53.28   & 64.96       & 58.40  & 63.75 & 60.41    & 51.73 & 57.19 &  58.59 & 55.81   \\
Diff. Wording & 25.39 & 53.55 & 59.04 &  54.64  & 47.40  & 60.56 & 45.55 & 57.76  & 51.65 & 58.44    & 56.68 & 56.18            \\ \midrule
+Definition & 51.55 & 59.14 & 59.03 & 58.79 & 65.36  & 63.84 & 64.64 & 60.99  & 52.66 & 53.55 & 57.42 & 50.35 \\
Diff. Wording & 58.64 & 58.16 & 63.56 & 59.44 & 57.03 & 62.63 & 59.02 & 61.07 & 50.80 & 54.53 & 57.06 & 49.71\\
\bottomrule
\end{tabular}
\caption{Comparative performance of GPT-3.5 Turbo, GPT-4 and Llama3 in Zero-Shot settings with varied prompt wordings.}
\label{differentwording}
\end{table*}

\paragraph{Relevance of Examples}
We further investigate how the relevance of examples to the test headlines, when available a priori, influences framing detection. We consider four settings:

\begin{itemize}
    \item \textbf{Focused in-domain}: In this setting, all the eight examples are relevant to gun violence, with all four framed headlines addressing a single aspect: health. 

    \item \textbf{Varied in-Domain}: In this setting framed examples cover four diverse aspects of gun violence\footnote{I.e., politics, public/school safety, race/ethnicity, and social/culture.} to test the model's adaptability to a range of in-domain cues.

\item \textbf{Cross-Domain}: To evaluate the model's performance on topics not known in advance, we use examples from completely different domains, such as immigration.

\item \textbf{Mixed Domain}: Combining in-domain and cross-domain examples, this scenario includes two framed instances related to gun violence and two from unrelated areas.

\end{itemize}

\subsection{Explanation-Based Prompting}
To further analyse model behaviour, we revisit both zero-shot and few-shot settings by requiring models to provide an explicit rationale for their predictions. Specifically, we append the instruction ``then give an explanation for your response'' to the original prompts, prompting models to justify their framing decisions alongside label predictions.

\section{Results and Analysis}
\label{sect:results}

We organise our analysis around four central dimensions highlighted in the Introduction: reliability under prompt variation, generalisation under domain shift, systematic failure modes, and cross-model agreement. In particular, we examine how prompting strategies affect prediction stability, how models generalise beyond the gun-violence domain, where systematic errors arise, and whether model agreement can provide insight into annotation quality.

\begin{table*}[!htbp]
    \centering
    \footnotesize
    \begin{tabular}{l|cc|cc|cc|cc} \toprule
    & \multicolumn{2}{c|}{\bfseries GPT-3.5-Turbo} & \multicolumn{2}{c}{\bfseries GPT-4} & \multicolumn{2}{c}{\bfseries Llama3 8B} & \multicolumn{2}{c}{\bfseries FLAN-T5 Large}\\
    & & & & & & & \multicolumn{2}{c}{\bfseries fine-tuned} \\
     & F$_1$ & Acc. & F$_1$ & Acc. & F$_1$ & Acc. & F$_1$ & Acc.\\ \midrule
     GVFC & 67.39 & 65.06 & 69.08 & 67.18 & 60.98 & 62.93 & \textbf{76.33} & \textbf{78.83} \\
     ITW & 75.68 & 71.34 & \textbf{82.49} & \textbf{80.25} & 76.14 & 73.25 & 69.18 & 68.78 \\
          \bottomrule
    \end{tabular}
    \caption{Performance on GVFC test set (FS varied in domain) and ITW dataset (FS cross-domain).}
    \label{tab:appflanfine}
\end{table*}

\subsection{Reliability and Prompt Sensitivity}

We first examine the stability of model predictions under different prompt formulations. Table~\ref{tab:gpt} reports performance under standard zero-shot and few-shot prompting with and without definitions, while Table~\ref{differentwording} isolates the effect of alternative prompt wording in zero-shot settings.

Together, these tables reveal substantial sensitivity to prompt design. For example, as shown in Table~\ref{differentwording}, GPT-4’s F$_1$ score drops from 64.96 in the standard zero-shot setting without a definition to 47.40 under alternative wording. A similar pattern is visible in Table~\ref{tab:gpt}, where zero-shot performance varies considerably across prompt variants.

Including an explicit framing definition reduces this variability. In both Table~\ref{differentwording} and Table~\ref{tab:gpt}, GPT-4’s F$_1$ scores remain more stable when a definition is provided. Explanation-based prompting further improves consistency, as reflected in reduced variance across explainable and non-explainable conditions. In contrast, Llama~3 exhibits comparatively smaller fluctuations across wording variants, suggesting greater robustness to prompt variation.

\subsection{Generalisation under Domain Shift}

We next investigate how well models generalise across domains and supervision regimes. We first analyse few-shot performance on the GVFC dataset, where topical information is available. Results are reported in Table~\ref{tab:gpt}. GPT-4 and Llama~3 achieve their highest F$_1$ scores in the 8-varied in-domain setting (68.51 and 58.58, respectively), indicating that diverse in-domain examples support framing detection when domain cues are accessible.

To evaluate performance under more realistic conditions, where headline topics are varied and unknown in advance, we construct a new In-the-Wild (ITW) dataset composed of real-world news headlines covering diverse subjects. Framed headlines are collected from a website dedicated to highlighting news framing\footnote{\url{https://newsframes.wordpress.com/category/headlines/}}, while non-framed instances are sampled from mainstream outlets, including the BBC, \emph{The Guardian}, and \emph{Daily Mail}, and selected to match framed headlines in terms of publication period and topical diversity. This sampling strategy aims to reduce potential source and temporal biases.
The resulting dataset contains 157 headlines spanning topics such as weather, public health, migration, and European affairs, including 83 framed and 74 non-framed instances, corresponding to a majority baseline accuracy of 52.86\%. All headlines were annotated by a domain expert following the framing definition introduced in Section~\ref{sec:intro}.

Evaluation results on the ITW dataset are reported in Table~\ref{tab:inthewild}. In the few-shot cross-domain setting, GPT-4 achieves its highest performance, reaching 80.25\% accuracy, demonstrating that instruction-following LLMs can generalise to heterogeneous real-world data when appropriately prompted. Nevertheless, performance remains more variable than in-domain evaluations, indicating persistent challenges under domain shift.

\begin{table*}[!htbp]
    \centering
    \footnotesize
    \begin{tabular}{l|cc|cc|cc} \toprule
    & \multicolumn{2}{c|}{\bfseries GPT-3.5 Turbo} & \multicolumn{2}{c|}{\bfseries GPT-4} & \multicolumn{2}{c}{\bfseries Llama3 8B}\\
     & F$_1$ & Acc. & F$_1$ & Acc. & F$_1$ & Acc.  \\ \midrule
     ZS + definition (S) & 74.16 & 70.70 & 78.61 & 72.61 & 60.34 & 54.78\\
     ZS + definition (E) & 74.15 & 66.24 & 78.00 & 71.97 & 63.27 & 54.14 \\
     FS8 cross-domain (S) & 67.50 & 66.88 &  81.11 & 78.34 & 69.33 & 70.70 \\
     FS8 cross-domain (E) & 75.68 & 71.34 & \bfseries 82.49 & \bfseries 80.25 & 76.14 & 73.25\\ 
          \bottomrule
    \end{tabular}
    \caption{Performance on the ITW Dataset, using the settings for Zero-Shot+Definition, Few-shot 8 examples cross domain. (S) stands for the standard setting, while (E) stands for the explainable setting. }
    \label{tab:inthewild}
\end{table*}

We further contrast pre-trained and fine-tuned models using this evaluation set. While fine-tuned FLAN-T5 Large performs strongly on GVFC, its performance deteriorates substantially on ITW, falling behind pre-trained GPT and Llama~3 models, as shown in Table \ref{tab:appflanfine}. This contrast highlights the limited transferability of task-specific fine-tuning and supports the use of pre-trained LLMs in low-resource, cross-domain settings.\footnote{The full results of FLAN-T5 models are available in appendix \ref{appendixa} and \ref{appendixflan}.}

\begin{table*}[!htb]
    \centering
    \footnotesize
    \begin{tabular}{l|cc|cc|cc} \toprule
     & \multicolumn{2}{c|}{\bfseries Llama3 8B} & \multicolumn{2}{c|}{\bfseries GPT-3.5} & \multicolumn{2}{c}{\bfseries GPT-4}\\
       &  F$_1$ & Acc. &  F$_1$ & Acc. &  F$_1$ & Acc.\\\midrule
      Clear & 76.60 & 75.00 &71.43 & 70.15 & 75.81 & 77.61 \\
        Contested & 21.31 & 27.82 & 41.43 & 38.81 & 3.60 & 20.15  \\ 
          \bottomrule
    \end{tabular}
    \caption{Comparative performance of the GPT and Llama3 models on the clear and contested framing subsets using the models' best performing settings.}
    \label{tab:SOevaluation}
\end{table*}

\begin{table*}[!htbp]
    \centering
    
    \resizebox{\textwidth}{!}{%
    \begin{tabular}{l} \toprule
       \textbf{Headline:} This list of lives lost in mass shootings grows longer. Here are their names.\\
        \textbf{Gold:} Not Framed, \textbf{GPT-4:} Framed, \textbf{GPT-3.5:} Not Framed, \textbf{Llama3:} Not Framed\\
        \textbf{GPT-4 Explanation:} This claim is framed because it specifically focuses on the tragic aspect of lives lost in mass shootings,\\evoking empathy and sadness. While it's factual that people have died, the phrasing and focus frame the issue in a particular \\emotional light.\\
        \textbf{GPT-3.5 Explanation:} This claim is not framed. It simply states that the list of lives lost in mass shootings is growing longer\\and presents the names of the victims. It does not provide any particular perspective or agenda.\\
        \textbf{Llama3 Explanation:} The headline does not explicitly promote a particular perspective or interpretation of the issue. It simply\\ presents the information in a neutral manner, without highlighting any specific aspect of the issue or promoting a \\particular agenda.\\
        \midrule
       
        \textbf{Headline:} 'Devastated' Tamera Mowry Confirms Niece Alaina Housley Died In Thousand Oaks Shooting\\
        \textbf{Gold:} Not Framed, \textbf{GPT-4:} Framed, \textbf{GPT-3.5:} Not Framed, \textbf{Llama3:} Not Framed\\
        \textbf{GPT-4 Explanation:} The claim is framed as it emphasizes the emotional state of Tamera Mowry (being 'devastated') while\\confirming the death of her niece in the shooting. This can elicit an emotional response from the audience and influence \\their perception of the issue.\\ 
        \textbf{GPT-3.5 Explanation:} The claim is not framed because it is simply reporting the confirmation of Tamera Mowry about the\\ death of her niece in the Thousand Oaks shooting. It does not select any specific aspects or favor a certain interpretation \\of the issue.\\
        \textbf{Llama3 Explanation:} This headline does not frame the issue by selecting certain aspects of the event, but rather \\reports the fact in a neutral way. It does not use emotive language, make value judgments, or highlight specific aspects \\of the event, but rather presents the information in a straightforward manner.\\ 
        \bottomrule
    \end{tabular}
    }
    \caption{Examples of the impact of emotional language on GPT-4.}
    \label{tab:examples}
\end{table*}
\subsection{Systematic Failure Modes}

Beyond aggregate performance, we examine recurring error patterns that limit model reliability. As illustrated by examples in Table~\ref{tab:examples}, GPT-4 frequently misclassifies emotionally charged headlines as framed, contributing to systematic false positives. This behaviour suggests that emotional intensity is often treated as a proxy for framing, even when no strategic framing is present.

To further analyse model behaviour on ambiguous instances, we conduct a manual review of 1,300 headlines sampled from the GVFC dataset. This analysis reveals discrepancies in the original annotations, including potentially incorrect labels and cases in which multiple framing interpretations are plausible.
Based on this review, we identify 134 \emph{contested} instances. These correspond to headlines whose framing status cannot be determined unambiguously or for which reasonable alternative interpretations exist. Examples include ``Live: Trump visits Pittsburgh after synagogue shooting'' and ``Shopify bans sale of certain firearms, accessories'', both annotated as framed in GVFC despite the absence of clearly identifiable framing strategies. At the same time, not all contested cases reflect erroneous annotations. Some headlines, such as ``Thousands gather to honor victims of the mass shooting with tears, candlelight, and song'', present more nuanced challenges in framing detection, where emotional content and descriptive language complicate interpretation. Within this subset, 63.4\% of headlines are annotated as framed.

For comparison, we also construct a \emph{clear} subset of 134 headlines for which the presence or absence of framing is readily identifiable based on the annotation guidelines. Representative examples include ``Two dead including shooter at Florida yoga studio'' (non-framed) and ``Parkland school shooter blames massacre on a `demon' voice'' (framed), both of which exhibit unambiguous framing status and align with their original annotations.

These two subsets enable a controlled analysis of model performance under conditions of low and high interpretive ambiguity.

Performance on these subsets is reported in Table~\ref{tab:SOevaluation}. While all models perform well on the clear subset, performance deteriorates sharply on contested cases, with GPT-4’s F$_1$ score dropping to 3.60. Similar degradation is observed for GPT-3.5 Turbo and Llama~3. These results indicate that current LLMs struggle when framing judgments require nuanced interpretation and contextual reasoning, even when they perform reliably on unambiguous instances.

\begin{table}[!htb]
    \centering
    \footnotesize
    \begin{tabular}{l|cc|cc} \toprule
    & \multicolumn{2}{c}{\bfseries Clear} & \multicolumn{2}{c}{\bfseries Contested}\\
     & Broad & Strict & Broad & Strict\\ \midrule
     Agr. & 100.00 &  54.55 & 100.00 & 46.62\\
     Agr. GL & 80.30  & 48.48 & 21.80 & 12.90 \\
          \bottomrule
    \end{tabular}
    \caption{Clear vs. Contested Annotations: Percentages of agreement between the predictions of GPT-3.5, GPT-4, and Llama, as well as their alignment with the gold label. "Broad Agreement" includes cases where at least two out of three models (2/3 or 3/3) agree on a prediction and match the Gold Label. "Strict Agreement" refers to cases where all three models (3/3) agree, and their prediction matches the gold label. The "Agreement" row indicates the percentage of headlines where at least two models agreed on a prediction, while the "Agreement GL" row shows the percentage of these agreed predictions that align with the gold label.}
    \label{tab:sopercentage}
\end{table}

\subsection{Cross-Model Agreement and Annotation Quality}

Finally, we examine whether patterns of agreement across models can provide insight into annotation quality. To this end, we analyse how often GPT-4, GPT-3.5 Turbo, and Llama~3 produce identical predictions on the clear and contested subsets. Agreement statistics are reported in Table~\ref{tab:sopercentage}.

On the contested subset, the three models reach unanimous agreement in 46.62\% of cases. However, only 12.90\% of these agreed-upon predictions match the original gold labels. In contrast, agreement on the clear subset is substantially higher and more frequently aligned with the annotations. These results indicate that strong cross-model consensus on contested cases often reflects disagreement with the existing labels rather than model error alone.

For example, all three models classify the headline ``Muslim Americans raise more than \$200,000 for those affected by Pittsburgh synagogue shooting'' as not framed, despite its annotation as framed in GVFC. Manual inspection suggests that this headline does not exhibit an obvious framing strategy, supporting the models’ shared interpretation.

Having manually reviewed the contested subset, we find that many instances with strong model agreement correspond to genuinely ambiguous or potentially problematic annotations. This suggests that cross-model consensus, particularly when it diverges from existing labels, can serve as a useful signal for identifying instances that merit further review.
Overall, these findings indicate that ensembles of LLMs may provide a practical tool for flagging potential annotation inaccuracies in existing or newly constructed datasets, especially in low-resource settings where large-scale re-annotation is infeasible.

\section{Conclusions}
In this work, we conducted a systematic investigation of instruction-following large language models for detecting framing in news headlines. Through extensive experiments across zero-shot, few-shot, and explanation-based settings, as well as in-domain and out-of-domain evaluations, we assessed model reliability, generalisation, and failure modes in a task characterised by inherent ambiguity and limited supervision.
Our results show that pre-trained LLMs, particularly GPT-4, can achieve strong performance when supported by carefully designed prompts and diverse few-shot examples. Explanation-based prompting improves prediction stability, while heterogeneous in-domain examples substantially enhance performance when topical information is available. At the same time, we identify persistent limitations, most notably a systematic tendency to conflate emotional language with framing and a pronounced performance collapse on contested cases requiring nuanced interpretation.
We further demonstrate that domain-specific fine-tuning, although effective in controlled settings, does not reliably transfer to heterogeneous real-world data. In contrast, pre-trained models exhibit greater adaptability under domain shift, highlighting their practical value for framing analysis in low-resource and rapidly evolving media environments.
Beyond model performance, our study highlights the importance of annotation quality in framing research. By analysing patterns of agreement across multiple models, we show that strong cross-model consensus can serve as a useful signal for identifying ambiguous or potentially problematic annotations. This provides a scalable mechanism for dataset auditing in contexts where expert annotation is costly or scarce.
Taken together, our findings underscore both the promise and the current limitations of LLM-based framing detection. While these models offer a flexible and scalable alternative to traditional supervised approaches, their reliability remains constrained by prompt sensitivity, annotation uncertainty, and the inherent complexity of framing. Our analysis and the datasets introduced in this work provide a foundation for future research in this direction.

\section{Limitations}

The findings of this study have to be seen in light of some limitations.
A significant constraint in the field of framing detection is the scarcity of expert-annotated datasets, which are not always publicly available. Even when such datasets are accessible, they often focus exclusively on framed data without including a balanced mix of framed and non-framed content.

Additionally, our evaluation focuses solely on English language content, leaving space for further investigation on other languages to explore our findings’ applicability to non-English contexts. This limitation suggests a need for further investigation into the performance of LLMs across different languages and cultural contexts to fully assess the potential use of these models in social science research for detecting framing and analysing media narratives.

Two of the five models evaluated in our work are accessible
only through OpenAI’s API, which is closed-source and subject to changes over time. This could affect the reproducibility of our results with newer versions of API, and they may have their own limitations. Therefore, focusing on improving open-source models emerges as a critical pathway forward, ensuring broader accessibility and reproducibility in research.

Finally, our binary framing classification (framed vs. not framed) simplifies a complex linguistic phenomenon. While framing often operates on a spectrum, treating it as a binary classification helps establish clearer methodological boundaries in framing detection studies. By enforcing a strict distinction between framed and non-framed content, this approach minimizes interpretative ambiguity, making it easier to compare framing patterns. Furthermore, a binary classification framework aligns with prior computational studies on media bias, allowing for more direct comparisons with existing research and ensuring reproducibility across different framing detection methodologies, and can aid in real-world applications, such as automatically flagging potentially framed news for further review. 

\section{Acknowledgements}

This work was supported by the Centre for Doctoral Training in Speech and Language Technologies (SLT) and their Applications funded by UK Research and Innovation [grant number EP/S023062/1].

\section{Bibliographical References}\label{sec:reference}

\bibliographystyle{lrec2026-natbib}
\bibliography{lrec2026-example}

\appendix
\section*{Appendix}
\section{FLAN-T5 Performance Lag}
\label{appendixa}
Table \ref{tab:flant5} shows the results of FLAN-T5 without any specific fine-tuning.

\begin{table*}[hbt!]
    \centering
    \footnotesize
    \resizebox{\textwidth}{!}{%
    \begin{tabular}{ll|cccc|cccc|cccc} \toprule
     & & \multicolumn{4}{c|}{FLAN-T5 Small} & \multicolumn{4}{c|}{FLAN-T5 Base} & \multicolumn{4}{c}{FLAN-T5 Large} \\
      & & & & \multicolumn{2}{c|}{Explainable} & & & \multicolumn{2}{c|}{Explainable} & & & \multicolumn{2}{c}{Explainable} \\
      & & F$_1$ & Acc. & F$_1$ & Acc. &  F$_1$ & Acc. &  F$_1$ & Acc. &  F$_1$ & Acc. &  F$_1$ & Acc.\\ \midrule
       \multirow{2}{*}{ZS}  & No Definition  & 0.46 & 49.79 & \bfseries 17.96 & 49.21 & 0 & 50.10 & \bfseries 25.86 & \bfseries 52.74 & 34.33 & 46.19 & 51.01 & \bfseries 49.06\\
         & +Definition & 0.15 & 50.13 & 1.21 & 50.10 & 0 & 50.10 & 0.61 & 50.06 & \bfseries 42.27 & 48.06 & 37.63 & 43.96 \\ 
         \midrule
         \multirow{5}{*}{FS}  
         & 2 examples & 0.31 & 50.06 & 0.31 & 50.06 & 0 & 50.10 & 3.12 & 49.98 & 5.74 & 49.64 & 8.32 & 49.64 \\
         & 8 (focused in-domain) & 0.15 & 50.17 & 3.84 & 50.10 & 0 & 50.13 & 4.83 & 50.17 & 16.78 & 47.53 & 36.85 & 44.69 \\
         & 8 (varied in-domain) & 0 & 50.13 & 0.15 & 49.94 & 0 & 50.13 & 1.66 & 49.94 & 22.78 & 45.96 & 35.03 & 45.27 \\
         & 8 (cross domain) & \bfseries 5.30 & 49.37 & 2.95 & 49.60 & 0 & 50.10 & 4.99 & 50.33 & 40.23 & 45.11 & \bfseries 51.64 & 46.88 \\
         & 8 (mixed domain) & 0.31 & \bfseries 50.21 & 1.51 & 50.02 & 0 & 50.10 & 3.84 & 50.13 & 34.39 & 44.58 & 44.05 & 43.43 \\
          \bottomrule
    \end{tabular}
    }
    \caption{Comparative performance of FLAN-T5 models using different prompt configurations.}
    \label{tab:flant5}
\end{table*}

\section{Fine-tuned FLAN-T5 and Hyperparameters}
\label{appendixflan}
Table \ref{tab:appflanfine2} presents the detailed performance comparison of fine-tuned FLAN-T5 small, base and large on the GVFC test set.
The FLAN-T5 models (small, base, and large) were fine-tuned using the following hyperparameters: All models employ a learning rate of 5e-05, with a maximum input length of 70 tokens, a maximum label length of 4 tokens, and a batch size of 64. The number of epochs varies with 10 for the small, 40 for the base, and 17 for the large model, each within a 50-epoch limit and with early stopping set at 5 epochs to prevent overfitting.
We used one A100 80G GPU, requiring less than 15 minutes of GPU time per model.

\begin{table*} [hbt!]
    \centering
    \footnotesize
    \begin{tabular}{l|cccc|cccc|cccc} \toprule
    & \multicolumn{4}{c|}{\bfseries FLAN-T5 Small} & \multicolumn{4}{c}{\bfseries FLAN-T5 Base} & \multicolumn{4}{c}{\bfseries FLAN-T5 Large}\\
    & \multicolumn{4}{c}{\bfseries fine-tuned} & \multicolumn{4}{c}{\bfseries fine-tuned} & \multicolumn{4}{c}{\bfseries fine-tuned} \\
     & F$_1$ & Acc. & Prec. & Rec. & F$_1$ & Acc. & Prec. & Rec. & F$_1$ & Acc. & Prec. & Rec.\\ \midrule
     GVFC & 0 & 52.29 & 0 & 0 & 79.77 & 79.50 & 74.82 & 85.42 & 76.33 & 78.83 & 80.63 & 72.46\\
     ITW & 2.38 & 47.70 & 100 & 1.20 & 63.87 & 56.05 & 56.48 & 73.49 & 69.18 & 68.78 & 72.36 & 66.26\\
          \bottomrule
    \end{tabular}
    \caption{Performance of fine-tuned FLAN-T5 Small, Base and Large on the GVFC test set and ITW dataset.}
    \label{tab:appflanfine2}
\end{table*}

\end{document}